%% file: main.tex
\documentclass[10pt,letterpaper]{article}

\usepackage[T1]{fontenc}
\usepackage[utf8]{inputenc}
\usepackage{lmodern}
\usepackage[letterpaper,margin=1in]{geometry}
\usepackage{amsmath,amssymb}
\usepackage{graphicx}
\usepackage{array,booktabs,multirow,tabularx}
\usepackage{url}
\usepackage[font=normalsize,labelfont=bf]{caption}

\graphicspath{{figures/}}
\makeatletter
\renewcommand\section{\@startsection{section}{1}{\z@}%
  {-2.2ex \@plus -0.6ex \@minus -0.2ex}%
  {0.9ex \@plus 0.2ex}%
  {\normalfont\large\bfseries}}
\renewcommand\subsection{\@startsection{subsection}{2}{\z@}%
  {-1.7ex \@plus -0.4ex \@minus -0.2ex}%
  {0.6ex \@plus 0.2ex}%
  {\normalfont\normalsize\itshape}}
\makeatother

\begin{document}

\twocolumn[
\begin{center}
  {\large\bfseries Trustworthy Medical Segmentation: Uncertainty-Aware U-Net
  Evaluation Under Clinical Image Degradation\par}
  \vspace{0.8em}
  \begin{minipage}{0.47\textwidth}
    \centering
    Pranav Kaliaperumal\\
    \textit{Department of Computer Science}\\
    \textit{University of Colorado Denver}\\
    Aurora, CO, USA\\
    \url{pranav.kaliaperumal@ucdenver.edu}
  \end{minipage}\hfill
  \begin{minipage}{0.47\textwidth}
    \centering
    Manisha Kaliaperumal\\
    \textit{Department of Biology (Pre-Medical)}\\
    \textit{Creighton University}\\
    Omaha, NE, USA\\
    \url{manishakaliaperumal@creighton.edu}
  \end{minipage}
\end{center}
\vspace{0.8em}

\noindent\textbf{\textit{Abstract}---}Medical image segmentation models often report high benchmark accuracy under ideal imaging conditions, yet their failures under clinical degradation can be quiet: sensor noise, patient motion, low-resolution acquisition, and contrast variability may all alter model behavior without producing an obvious warning. We present a reproducible framework for evaluating uncertainty-aware segmentation under controlled clinical degradation. Our experiments use a synthetic multimodal brain tumor MRI cohort generated with a biophysical phantom simulator that follows the BraTS protocol. We train U-Net and Attention U-Net baselines for multi-class tumor sub-region segmentation and augment both models with Monte Carlo dropout to estimate per-voxel uncertainty. Across eight clinically motivated corruption types at five severity levels, we measure segmentation accuracy, calibration, failure detection, and selective prediction coverage. On clean data, Attention U-Net achieves a whole-tumor Dice of 0.990; under severe Gaussian noise, its performance falls to 0.089. Predictive uncertainty rises with degradation and tracks segmentation error (Pearson $r=0.53$ under severity-3 Gaussian noise), allowing us to flag failures with an AUROC of 0.843. These results argue for uncertainty-aware inference as a practical safety layer in physician-in-the-loop radiology workflows. We release the code, trained models, and evaluation protocol to support direct reproduction.

\vspace{0.5em}
\noindent\textit{Index Terms}---Medical image segmentation, uncertainty quantification, Monte Carlo dropout, robustness evaluation, brain tumor segmentation, attention mechanism, clinical image degradation, synthetic phantom.
\vspace{1em}
]

\input{sections/introduction}
\input{sections/related_work}
\input{sections/methods}
\input{sections/experimental_data}
\input{sections/experimental_setup}
\input{sections/results}
\input{sections/discussion}
\input{sections/conclusion}

\bibliographystyle{unsrt}
\bibliography{references}

\end{document}

%% file: sections/introduction.tex
\section{Introduction}
\label{sec:introduction}

Deep learning-based medical image segmentation has moved quickly from a research curiosity to a clinical prototype. U-Net variants and their attention-augmented successors now reach near-radiologist performance on benchmark datasets for organ delineation, tumor boundary detection, and lesion segmentation \cite{ronneberger2015unet,oktay2018attention}. Yet the translation gap remains wide \cite{mehrtash2020confidence}. Clinical magnetic resonance imaging (MRI) differs from curated benchmark data in ways that matter: scans may contain thermal sensor noise from low-field scanners, motion artifacts from claustrophobic or pediatric patients, variable contrast from inconsistent gadolinium protocols, Gibbs ringing from accelerated acquisition, and partial-volume effects at tissue boundaries \cite{hendrycks2019benchmarking,kamnitsas2017efficient}. When a segmentation model meets these out-of-distribution conditions, it often produces a degraded mask without signaling reduced confidence. A neuro-oncologist using an automated tumor boundary for radiotherapy planning should not have to infer whether the algorithm processed a pristine 3T MRI or a motion-corrupted, low-dose acquisition from a rural clinic. We view this lack of calibrated trust as a central barrier to clinical adoption \cite{ghesu2019quantifying}.

Uncertainty quantification gives us a way to make that trust visible. Voxel-level uncertainty maps can identify regions where a segmentation is unreliable and direct expert review to the boundaries that need it most \cite{gal2016dropout}. Monte Carlo (MC) dropout, introduced by Gal and Ghahramani, is especially useful here because it turns a dropout-equipped neural network into an approximate Bayesian model without changing the architecture or requiring costly posterior inference \cite{gal2016dropout}. Multiple stochastic forward passes produce a predictive distribution. From that distribution, we can estimate both epistemic uncertainty, reflecting gaps in model knowledge, and aleatoric uncertainty, reflecting ambiguity in the image itself.

\subsection{Contributions}

We make six contributions:
\begin{enumerate}
    \item We introduce a reproducible robustness benchmark for medical image segmentation under eight clinically motivated corruption types at five severity levels, extending the ImageNet-C \cite{hendrycks2019benchmarking} idea to multimodal brain MRI.
    \item We generate a controlled synthetic cohort of 60 multimodal brain tumor MRI volumes with BraTS-convention labels, using an open-source biophysical phantom simulator so that the evaluation has no data-use restrictions.
    \item We compare U-Net and Attention U-Net on tumor sub-region segmentation and quantify how attention gating behaves under clean and degraded conditions.
    \item We add MC-dropout inference and calibration analysis to test whether predictive uncertainty identifies segmentation failures when corruptions become severe.
    \item We analyze selective prediction as a coverage-accuracy tradeoff, giving a tunable route for human-in-the-loop workflows in which uncertain predictions move to neuroradiologist review.
    \item We provide a one-command reproducibility pipeline, including phantom simulation, model training, corruption functions, uncertainty estimation, figure generation, and paper assembly.
\end{enumerate}

Accurate glioma segmentation matters because it turns a complex image into a quantitative description of tumor boundaries and subregions. Neuro-oncologists use these contours to estimate tumor burden, define surgical or radiotherapy targets, and track response or progression over time \cite{jin2020artificial}. Errors are not abstract. Under-segmentation can leave tumor outside a planned resection or radiation field; over-segmentation can expose healthy brain tissue to unnecessary treatment and increase the risk of neurological deficit \cite{wong2020comparing}. The quality of the mask therefore propagates into treatment efficacy, prognostic assessment, and patient outcomes.

%% file: sections/related_work.tex
\section{Related Work}
\label{sec:related-work}

\subsection{Medical Image Segmentation Architectures}

The U-Net architecture, introduced by Ronneberger et al. for biomedical image segmentation, remains the foundational encoder-decoder design for medical imaging \cite{ronneberger2015unet}. Its skip connections preserve spatial detail at multiple resolutions, making it particularly effective for fine boundary delineation. Subsequent variants have introduced attention mechanisms, nested skip pathways, and transformer-based encoders. Oktay et al. proposed Attention U-Net, which inserts attention gates at skip connections to suppress irrelevant background activations and highlight foreground saliency, achieving measurable improvements on abdominal CT and cardiac MRI \cite{oktay2018attention}. Attention mechanisms have since been shown to improve both accuracy and interpretability across diverse medical imaging tasks \cite{xie2023attention,gu2021canet}.

For brain tumor segmentation specifically, the BraTS challenge has driven architectural innovation for over a decade \cite{menze2015brats,bakas2017advancing,bakas2018identifying}. State-of-the-art methods now combine U-Net backbones with residual connections, deep supervision, and multi-scale feature fusion \cite{isensee2021nnunet}. Myronenko demonstrated that a well-regularized U-Net with deep supervision and region-specific loss weighting achieves competitive BraTS performance without complex architectural modifications \cite{myronenko2018brain}. More recently, nnU-Net established a self-configuring framework that automatically adapts preprocessing, network topology, and postprocessing to new datasets, becoming a de facto standard for medical segmentation benchmarking \cite{isensee2021nnunet}.

\subsection{Uncertainty Quantification in Medical Imaging}

Bayesian approaches to uncertainty have become increasingly important in medical imaging. Mehrtash et al. showed that MC-dropout uncertainty in prostate MRI segmentation correlates with inter-observer disagreement among radiologists, giving algorithmic uncertainty a clinical anchor \cite{mehrtash2020confidence}. Kwon et al. applied Bayesian neural networks to ischemic stroke lesion segmentation and showed that uncertainty-guided active learning can reduce annotation cost \cite{kwon2020uncertainty}. Jungo and Reyes evaluated uncertainty estimation across medical image segmentation tasks, finding that reliability determines whether uncertainty is useful at the bedside \cite{jungo2019assessing}. Carannante et al. reached a similar conclusion in a benchmark focused on trustworthy medical segmentation \cite{carannante2021trustworthy}.

The distinction between epistemic uncertainty, which can shrink with more data, and aleatoric uncertainty, which belongs to the imaging process itself, is especially relevant in medicine \cite{kendall2017uncertainties}. Epistemic uncertainty can flag rare tumor morphologies or unfamiliar acquisition protocols. Aleatoric uncertainty captures genuine ambiguity at boundaries such as the transition between necrotic and enhancing tumor. We quantify predictive uncertainty with MC dropout because it directly targets the out-of-distribution degradation scenarios most likely to cause clinical failure.

\subsection{Robustness and Corruption Benchmarking}

Hendrycks and Dietterich introduced ImageNet-C, a systematic corruption benchmark that exposed substantial robustness gaps in state-of-the-art classification models \cite{hendrycks2019benchmarking}. Medical imaging has adopted this style of stress testing more slowly. M{\aa}rtensson et al. evaluated a deep learning model on clinical out-of-distribution MRI across multiple cohorts and found major performance losses when scanners and protocols differed from the training data \cite{martensson2020reliability}. Panfilov et al. studied denoising convolutional networks as a route to more robust medical image diagnosis \cite{panfilov2019improving}. We extend this line of work with a corruption benchmark tailored to brain tumor segmentation and paired with uncertainty calibration analysis.

Uncertainty is not unique to algorithms: radiologists handle it in manual tumor segmentation through multi-rater annotation and consensus-building, where discrepancies are resolved through discussion and majority agreement. In datasets such as BraTS, ground truth labels are derived from the annotations of multiple experts fused into a single reference standard, implicitly encoding inter-observer variability \cite{menze2015brats}. This variability is persistent: differences in how radiologists interpret tumor boundaries---particularly between enhancing tumor, edema, and necrotic regions---introduce label noise that affects both model training and evaluation \cite{menze2015brats}. Prior studies have shown that inter-observer disagreement can be substantial in glioma segmentation, indicating that ``ground truth'' labels are better understood as approximations than as absolute references \cite{porz2014multimodal}.

%% file: sections/methods.tex
\section{Methods}
\label{sec:methods}

\subsection{Architecture Overview}

We evaluate two backbone architectures for brain tumor sub-region segmentation:

\emph{U-Net Baseline.} Our U-Net implementation follows the standard encoder-decoder design with four downsampling and four upsampling blocks connected by skip pathways. Each block contains two $3\times3$ convolutions with batch normalization and ReLU activation. For MC-dropout segmentation, we place 2D dropout layers ($p=0.1$) in the deep semantic blocks---the third encoder stage, fourth encoder stage, bottleneck, and first two decoder stages---while keeping low-level feature extraction deterministic. We activate dropout during training and again during MC-dropout inference. The encoder halves spatial resolution while doubling channels: $12\rightarrow24\rightarrow48\rightarrow96\rightarrow192$. The decoder reverses this pathway and concatenates skip connections at each level.

\emph{Attention U-Net.} Building on the U-Net baseline, we replace standard skip connections with attention gates following Oktay et al. \cite{oktay2018attention}. Each attention gate takes the gating signal from the coarser decoder feature map and the skip connection feature map from the encoder, producing a spatial attention coefficient:
\begin{equation}
q_{\mathrm{att}}^{l}
=\sigma_{1}\!\left(W_{x}^{T}x^{l}+W_{g}^{T}g\right),
\label{eq:attention-score}
\end{equation}
\begin{equation}
\alpha^{l}=\sigma_{2}\!\left(\psi^{T}q_{\mathrm{att}}^{l}+b_{\psi}\right),
\label{eq:attention-coefficient}
\end{equation}
\begin{equation}
\hat{x}_{i,l}=\alpha_{i,l}\cdot x_{i,l},
\label{eq:attention-feature}
\end{equation}
where $x^{l}$ is the skip connection feature, $g$ is the gating signal, and $\alpha$ is the attention coefficient. The suppressed features $\hat{x}$ are concatenated with the upsampled decoder features, reducing background noise propagation.

\subsection{Monte Carlo Dropout for Uncertainty Estimation}

Following Gal and Ghahramani \cite{gal2016dropout}, we approximate Bayesian inference by performing $T$ stochastic forward passes with dropout enabled at inference time. For each voxel $i$, we collect predictions $\{\hat{y}_{i}^{(t)}\}_{t=1}^{T}$ and compute the predictive mean
\begin{equation}
\bar{y}_{i}=\frac{1}{T}\sum_{t=1}^{T}\hat{y}_{i}^{(t)},
\label{eq:predictive-mean}
\end{equation}
the predictive uncertainty (entropy)
\begin{equation}
H[y_i]= -\sum_{c=1}^{C}\bar{y}_{i,c}\log \bar{y}_{i,c},
\label{eq:predictive-entropy}
\end{equation}
and the epistemic uncertainty (mutual information)
\begin{equation}
\begin{split}
\operatorname{MI}[y_i,\theta\mid x_i,D]
={}&H[y_i\mid x_i,D]\\
&-\mathbb{E}_{p(\theta\mid D)}\!\left[H[y_i\mid x_i,\theta]\right].
\end{split}
\label{eq:mutual-information}
\end{equation}
Although the framework defines both quantities, we use predictive entropy $H[y_i]$, averaged over all voxels in a slice, as our scalar uncertainty score. This score captures the total uncertainty a reviewer would see and act upon. We use $T=30$ forward passes to balance uncertainty quality against computational overhead. On our CPU-only evaluation hardware, a single deterministic forward pass over one $128\times128$ slice takes approximately 11 ms, and a full $T=30$ uncertainty estimate takes approximately 0.4 s per slice.

\subsection{Clinically Motivated Image Degradation}

We define eight corruption types simulating common clinical MRI conditions, summarized in Table~\ref{tab:corruptions}. Each corruption is applied at five severity levels (1--5), producing 40 distinct degradation conditions plus the clean baseline. Severity scales are calibrated on held-out data so that severity 3 represents moderate clinical degradation and severity 5 represents severe degradation rarely encountered clinically.

\begin{table*}[t]
\centering
\caption{Clinically Motivated Corruption Functions}
\label{tab:corruptions}
\begin{tabular}{@{}p{0.18\textwidth}p{0.27\textwidth}p{0.49\textwidth}@{}}
\hline
\textbf{Degradation} & \textbf{Clinical Origin} & \textbf{Implementation} \\
\hline
Gaussian Noise & Thermal sensor noise & Additive $\mathcal{N}(0,\sigma I)$, $\sigma=0.15$--$1.40$ \\
Motion Blur & Patient movement & Linear kernel convolution, 3--13 px \\
Defocus Blur & Off-resonance, shimming & Gaussian kernel, $\sigma=0.7$--$2.6$ px \\
Low Resolution & Accelerated scanning & $1.5\times$--$6\times$ downsample-upsample \\
Contrast Shift & Protocol variation & Gamma transform, $\gamma=0.8$--$0.22$ \\
Brightness Shift & Coil sensitivity & Additive intensity delta, 0.3--2.5 \\
Occlusion & Susceptibility artifact & 4--14 random zero patches, 12--30 px \\
JPEG Compression & PACS transmission & Quality factor 70--5 \\
\hline
\end{tabular}
\end{table*}

During clinical shadowing, we observed two of these degradation modes repeatedly. Motion blur, often caused by patient movement or physiological motion, was a common source of MRI degradation; radiologists addressed it with motion-correction sequences such as PROPELLER \cite{pipe1999propeller} or by repeating affected acquisitions. Susceptibility-related occlusions, especially near metallic implants, required changes such as echo-time optimization or metal artifact reduction sequences to preserve diagnostic utility.

\subsection{Training Protocol}

All models are trained with a combined cross-entropy and Dice loss:
\begin{equation}
\mathcal{L}=\lambda_{\mathrm{CE}}\mathcal{L}_{\mathrm{CE}}+\lambda_{\mathrm{Dice}}\mathcal{L}_{\mathrm{Dice}},
\label{eq:combined-loss}
\end{equation}
with $\lambda_{\mathrm{CE}}=\lambda_{\mathrm{Dice}}=0.5$. The multi-class soft Dice loss over $C$ tumor sub-regions is
\begin{equation}
\mathcal{L}_{\mathrm{Dice}}
=1-\frac{1}{C}\sum_{c=1}^{C}
\frac{2\sum_i y_{i,c}\hat{y}_{i,c}+\epsilon}
{\sum_i y_{i,c}+\sum_i\hat{y}_{i,c}+\epsilon}.
\label{eq:dice-loss}
\end{equation}
Training uses AdamW optimization with initial learning rate $10^{-3}$, cosine annealing schedule, weight decay $10^{-4}$, and early stopping with patience of 7 epochs based on validation whole-tumor Dice. Input slices are $128\times128$ and z-score normalized per modality within the head mask. Data augmentation includes random horizontal and vertical flipping, $90^{\circ}$ rotations, and rotation ($\pm15^{\circ}$).

\subsection{Evaluation Metrics}

Following community recommendations for image analysis validation \cite{maierhein2024metrics}, we evaluate segmentation quality using:
\begin{itemize}
    \item Dice Similarity Coefficient (DSC): $\mathrm{DSC}=2\lvert X\cap Y\rvert/(\lvert X\rvert+\lvert Y\rvert)$.
    \item Intersection over Union (IoU): $\mathrm{IoU}=\lvert X\cap Y\rvert/\lvert X\cup Y\rvert$.
    \item 95th-percentile Hausdorff Distance (HD95): robust maximum surface distance between prediction and ground truth, reported in pixels.
    \item Boundary Dice: Dice computed on boundary regions dilated by 2 pixels.
\end{itemize}

For uncertainty quality, we compute:
\begin{itemize}
    \item Uncertainty Calibration Error (UCE): alignment between predicted uncertainty and actual voxel error across quantile bins.
    \item Expected Calibration Error (ECE): standard confidence calibration across probability bins.
    \item Failure Detection AUROC: discrimination of failed slices (whole-tumor Dice $<0.65$) using the slice-level mean predictive entropy as the failure predictor.
    \item Uncertainty--Error Correlation: Pearson $r$ between slice-level mean predictive entropy and segmentation error ($1-\mathrm{Dice}$).
    \item Selective Prediction Coverage: fraction of predictions retained at varying uncertainty thresholds, with the Dice computed on retained predictions.
\end{itemize}

%% file: sections/experimental_data.tex
\section{Experimental Data}
\label{sec:experimental-data}

\subsection{Synthetic Multimodal Cohort}

We use a synthetic cohort of 60 brain tumor patient volumes generated by a purpose-built biophysical phantom simulator in the tradition of the BrainWeb simulated brain database. For each patient, the simulator constructs a three-dimensional head volume ($64\times128\times128$ voxels) containing cerebrospinal fluid, gray matter, white matter, and skull compartments. Smooth Gaussian random fields and elastic deformation perturb the tissue boundaries, giving the cohort controlled anatomical variability. We model tumors as nested compartments---an infiltrative peritumoral edema front, a gadolinium-enhancing rim, and a necrotic/non-enhancing core---using the BraTS label convention (NCR/NET label 1, ED label 2, ET label 4). We then synthesize four MRI modalities (T1, T1Gd, T2, and T2-FLAIR) from tissue- and compartment-specific intensity priors with coherent texture, a multiplicative low-frequency bias field, partial-volume smoothing, per-acquisition contrast and SNR variation, and additive sensor noise. Each modality is z-score normalized within the head mask.

Our contrast model follows clinical practice. Gadolinium is a paramagnetic MRI contrast agent that brightens regions where the blood-brain barrier is disrupted, a pattern often associated with active or aggressive tumor growth \cite{acr2024contrast}. This contrast helps clinicians distinguish tumor tissue from healthy brain, identify active growth, and evaluate disease spread, making contrast-enhanced MRI central to treatment planning and monitoring \cite{acs2025mri}. We simulate four modalities because each contributes different information: T1-weighted MRI shows basic anatomy; T1 with gadolinium highlights active tumor and abnormal vasculature; T2-weighted MRI is sensitive to fluid and peritumoral edema; and FLAIR suppresses normal CSF signal so abnormalities near fluid-filled spaces are easier to see \cite{acs2025mri}. Together, the modalities describe tumor size, structure, activity, and surrounding tissue change more completely than any single scan.

\begin{figure*}[t]
\centering
\includegraphics[width=\textwidth]{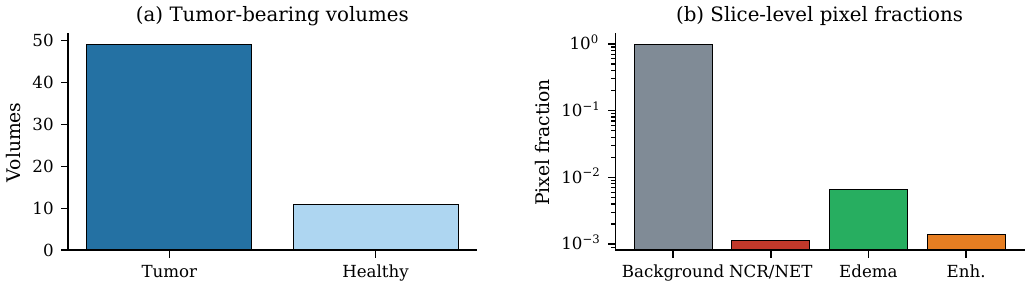}
\caption{(a) Volume-level tumor presence distribution and (b) slice-level pixel class distribution in the synthetic cohort. Background pixels dominate, creating severe class imbalance typical of medical segmentation tasks.}
\label{fig:dataset}
\end{figure*}

From the 60 volumes, we extract 2,400 axial slices (40 per volume), sampling tumor-bearing slices with 80\% probability. We split at the patient level: 42 volumes (1,680 slices) for training, 8 volumes (320 slices) for validation, and 10 volumes (400 slices) for testing. No patient appears in more than one split. Fig.~\ref{fig:dataset} summarizes the cohort: 49 of 60 volumes (82\%) contain tumor, 47\% of extracted slices are tumor-bearing, and background pixels dominate at the slice level (mean background ratio 0.99). This imbalance is typical of tumor segmentation and makes whole-image accuracy misleading unless region-specific metrics are reported.

\subsection{Clinical Workflow Context}

Segmentation models do not operate in isolation; they eventually meet physician documentation, structured reports, and electronic health records. The NBME clinical patient notes dataset \cite{kaggle2022nbme}, a public collection of de-identified physician notes with annotated clinical concepts, illustrates this multimodal setting. A brain tumor segmentation model could generate quantitative measurements---tumor volume, edema extent, and anatomical location---and insert them into a draft radiology report through structured reporting. Recent work suggests that incorporating AI-generated imaging measurements into radiology reports can improve reporting efficiency and support clinical decision-making by combining imaging outputs with patient history and pathology information \cite{tordjman2025llm}. Radiology natural language processing studies likewise show that linking imaging data with clinical narratives can support more comprehensive multimodal AI systems \cite{pons2016nlp}. We leave validation on real clinical scans, including BraTS \cite{menze2015brats,bakas2017advancing,bakas2018identifying}, to future work; here, the synthetic cohort gives us a controlled and restriction-free testbed.

%% file: sections/experimental_setup.tex
\section{Experimental Setup}
\label{sec:experimental-setup}

\subsection{Implementation Details}

We implement all experiments in PyTorch 2.8 and run them in a CPU-only environment (2 cores, 4 GB RAM). This hardware choice is deliberate: the full pipeline should be reproducible without specialized accelerators. Training takes approximately 11 minutes per model for up to 18 epochs with early stopping (18 epochs for U-Net, 18 for Attention U-Net). We use a batch size of 24 for both models. U-Net has 1.09M parameters, and Attention U-Net has 1.11M. The complete experimental suite---training both models, evaluating 40 corruption conditions, and computing uncertainty metrics---takes approximately 0.8 hours.

\subsection{Experimental Protocol}

We execute three experimental phases:
\begin{enumerate}
    \item Baseline training: train U-Net and Attention U-Net on clean training data for up to 18 epochs with early stopping.
    \item Corruption evaluation: evaluate both models on clean test data and all 40 corrupted variants (8 corruptions $\times$ 5 severities) on the full 400-slice test set.
    \item Uncertainty analysis: compute MC dropout uncertainty maps ($T=30$) for clean data and all eight corruptions at severity 3 on 120 tumor-bearing test slices, and evaluate calibration, failure detection, and selective prediction metrics.
\end{enumerate}

%% file: sections/results.tex
\section{Results}
\label{sec:results}

\subsection{Clean Test Set Performance}

Table~\ref{tab:clean-performance} reports clean test-set performance for both architectures across the standard BraTS region hierarchy. Attention U-Net improves every tumor sub-region relative to the standard U-Net, with the largest gain on enhancing tumor ($+0.48$ Dice points) and a whole-tumor gain of $+0.17$ Dice points. HD95 improves by 0.03 pixels, suggesting sharper boundary placement. The whole-tumor Dice values sit in the numerical range often reported for 2D U-Net-style glioma segmentation~\cite{isensee2021nnunet,myronenko2018brain}, but our synthetic cohort is easier and more controlled than real BraTS data. We therefore interpret the absolute values cautiously and focus on relative behavior under matched degradations.

\begin{table*}[!t]
\caption{Clean Test Set Segmentation Performance}
\label{tab:clean-performance}
\centering
\setlength{\tabcolsep}{4pt}
\begin{tabular*}{\textwidth}{@{\extracolsep{\fill}}lcccccccc@{}}
\hline
& \multicolumn{2}{c}{Enh. Tumor} & \multicolumn{2}{c}{Tumor Core} & \multicolumn{2}{c}{Whole Tumor} & & \\
\cline{2-3}\cline{4-5}\cline{6-7}
Model & Dice & IoU & Dice & IoU & Dice & IoU & HD95 (px) & Params \\
\hline
U-Net       & 0.9704 & 0.9507 & 0.9773 & 0.9630 & 0.9881 & 0.9775 & 0.07 & 1.09M \\
Attn U-Net  & 0.9751 & 0.9567 & 0.9820 & 0.9690 & 0.990  & 0.9805 & 0.05 & 1.11M \\
\hline
\end{tabular*}
\end{table*}

\subsection{Robustness Under Clinical Degradation}

\begin{figure*}[!t]
\centering
\includegraphics[width=\textwidth]{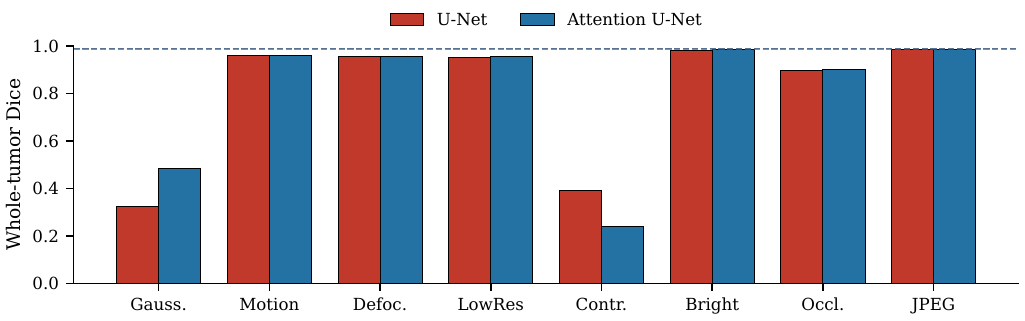}
\caption{Whole-tumor Dice coefficient under eight corruption types at severity level 3. Dashed lines show clean-data performance for each model.}
\label{fig:corruption-bars}
\end{figure*}

Table~\ref{tab:severity3} presents whole-tumor Dice under severity-3 corruption. Degradation is substantial, and the type of degradation matters (Fig.~\ref{fig:corruption-bars}). Contrast shift produces the largest drop (Attention U-Net: 75.6\%; U-Net: 60.2\%), followed by Gaussian noise. JPEG compression and brightness shift are better tolerated at this severity. Attention U-Net outperforms U-Net in 7 of the eight corruption types, but the advantage is not universal. Under heavy additive noise and extreme contrast shift, attention gating does not by itself guarantee robustness; it improves feature selection, not clinical safety.

\begin{table*}[!t]
\caption{Whole Tumor Dice Under Severity-3 Clinical Degradation}
\label{tab:severity3}
\centering
\setlength{\tabcolsep}{4pt}
\begin{tabular*}{\textwidth}{@{\extracolsep{\fill}}lccccccccc@{}}
\hline
Model & Clean & Gauss. & Motion & Defoc. & LowRes & Contr. & Bright & Occl. & JPEG \\
\hline
U-Net      & 0.9881 & 0.323 & 0.962 & 0.958 & 0.953 & 0.393 & 0.983 & 0.898 & 0.985 \\
Attn U-Net & 0.990  & 0.486 & 0.962 & 0.958 & 0.955 & 0.241 & 0.987 & 0.900 & 0.988 \\
\hline
\end{tabular*}
\end{table*}

\subsection{Severity Progression Analysis}

\begin{figure*}[!t]
\centering
\includegraphics[width=\textwidth]{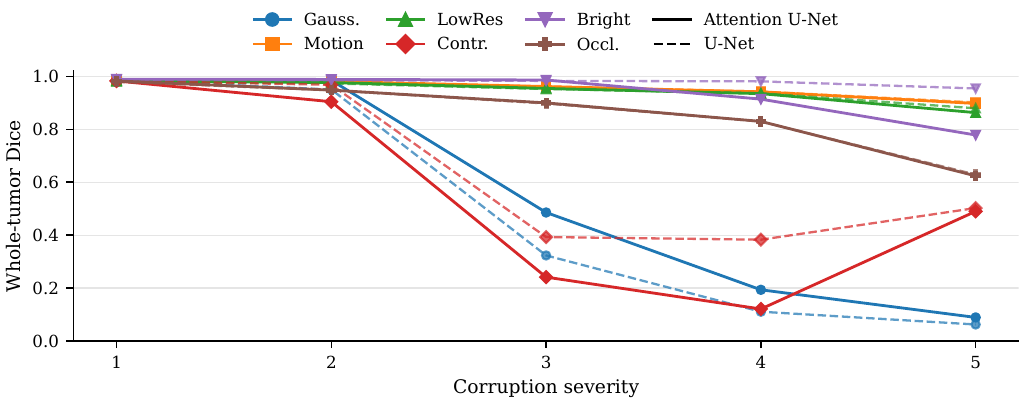}
\caption{Progressive performance degradation as corruption severity increases from 1 (mild) to 5 (severe). Solid lines: Attention U-Net; dashed lines: U-Net.}
\label{fig:severity}
\end{figure*}

Figure~\ref{fig:severity} shows progressive degradation across five severity levels, and Table~\ref{tab:gaussian-progression} gives the Gaussian-noise trajectory in detail. Attention U-Net tolerates mild Gaussian noise (whole-tumor Dice 0.986 at severity 2), but performance breaks sharply once noise reaches severity 3 and falls to 0.089 by severity 5. Other corruptions degrade more gradually. Motion blur, defocus blur, low resolution, occlusion, and JPEG compression preserve more spatial structure, so they tend to erode performance rather than collapse it. The sharper failures occur when noise or global intensity changes disrupt the contrast cues on which the convolutional filters rely.

\begin{table*}[!t]
\caption{Gaussian Noise Severity Progression (Whole Tumor Dice)}
\label{tab:gaussian-progression}
\centering
\begin{tabular}{lccccc}
\hline
Model & Sev 1 & Sev 2 & Sev 3 & Sev 4 & Sev 5 \\
\hline
U-Net      & 0.988 & 0.951 & 0.323 & 0.110 & 0.062 \\
Attn U-Net & 0.989 & 0.986 & 0.486 & 0.193 & 0.089 \\
\hline
\end{tabular}
\end{table*}

These degradation levels also need clinical interpretation. In radiotherapy planning, useful automated brain tumor segmentations generally require Dice scores above approximately 0.75--0.86 to maintain spatial agreement with expert contours, especially when treatment margins approach critical brain structures~\cite{wong2020comparing,bakx2023clinical}. No single threshold applies to every case, and radiation oncologists or neuroradiologists still review and correct automated contours. In our experiments, most severity-3 corruptions remain within or near this band, but additive sensor noise and strong contrast shift fall outside it: Attention U-Net reaches 0.486 under severity-3 Gaussian noise and 0.089 by severity 5. This is precisely the regime where explicit failure detection becomes necessary.

\subsection{Uncertainty Calibration Analysis}

\begin{figure*}[!t]
\centering
\includegraphics[width=\textwidth]{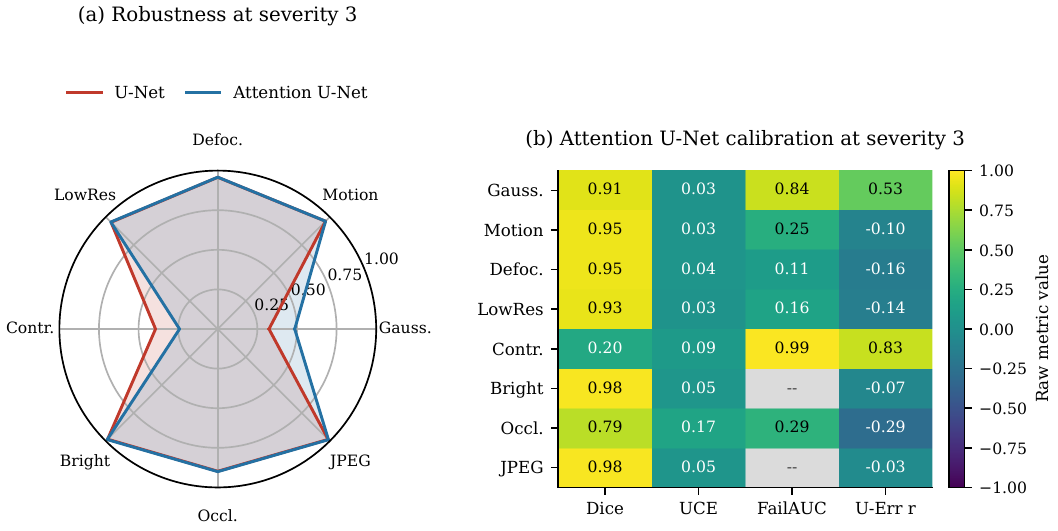}
\caption{(a) Radar chart comparing model robustness across the corruption spectrum at severity 3. (b) Heatmap of segmentation and calibration metrics (Dice, UCE, failure AUROC, uncertainty--error correlation) across corruption types for Attention U-Net at severity 3.}
\label{fig:radar-heatmap}
\end{figure*}

\begin{figure*}[!t]
\centering
\includegraphics[width=\textwidth]{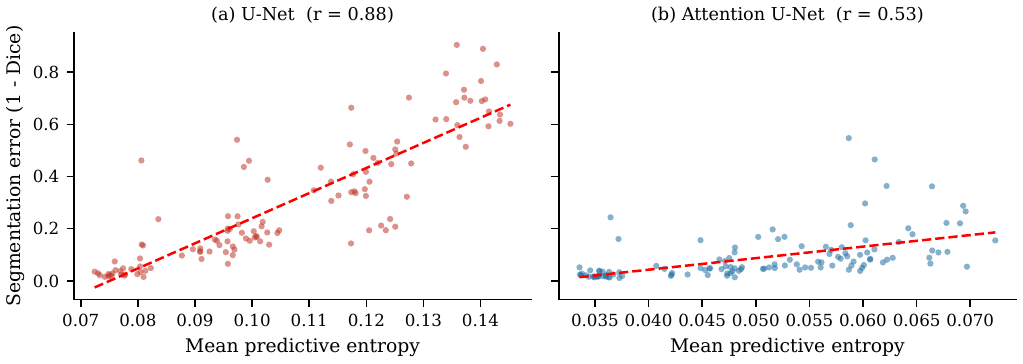}
\caption{Scatter plots of slice-level mean predictive entropy versus segmentation error for (a) U-Net ($r=0.88$) and (b) Attention U-Net ($r=0.53$) under Gaussian noise at severity 3. Each point represents one test slice; the red dashed line shows the linear regression fit.}
\label{fig:uncertainty-scatter}
\end{figure*}

\begin{figure*}[!t]
\centering
\includegraphics[width=0.96\textwidth]{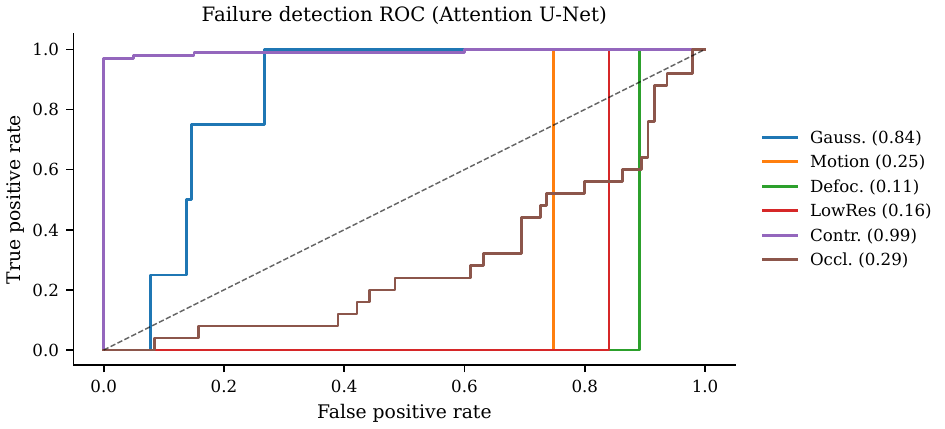}
\caption{Failure detection ROC curves using slice-level mean predictive entropy as the failure predictor (Attention U-Net). All corruption conditions are evaluated at severity 3; conditions with no failed slices are omitted because an ROC cannot be defined for them.}
\label{fig:failure-roc}
\end{figure*}

Table~\ref{tab:uncertainty-calibration} reports Attention U-Net uncertainty calibration under representative corruptions, and Fig.~\ref{fig:radar-heatmap} summarizes both robustness and calibration. The stochastic predictive mean is often more stable than a single deterministic pass: under severity-3 Gaussian noise, it attains a whole-tumor Dice of 0.911 compared with 0.486 for deterministic inference. This gap reflects the regularizing effect of model averaging when the input is noisy. Three findings stand out:

\begin{enumerate}
\item Uncertainty rises with degradation and detects failure: under severity-3 Gaussian noise, predictive entropy separates failed from successful segmentations with AUROC 0.843. On clean data, the model is uniformly confident and correct (minimum whole-tumor Dice 0.887, with 0 of 120 slices below the 0.65 failure threshold).
\item Uncertainty correlates with error: the slice-level uncertainty--error correlation reaches $r=0.53$ under severe Gaussian noise (Fig.~\ref{fig:uncertainty-scatter}), meaning that a substantial fraction of variance in segmentation error is predictable from uncertainty alone.
\item Within the severity-3 conditions, Contrast shift failures are best detected (AUROC $=0.992$; Fig.~\ref{fig:failure-roc}), while milder corruptions produce too few failures for reliable discrimination. Occlusion remains an instructive intermediate case: failures can occur, but entropy may rank them poorly (AUROC 0.293), suggesting that the uncertainty signal is strongest for global intensity degradations and less dependable for spatially localized artifacts.
\end{enumerate}

\begin{table*}[!t]
\caption{Uncertainty Calibration Metrics Under Corruption (Attention U-Net, Severity 3)}
\label{tab:uncertainty-calibration}
\centering
\begin{tabular}{lccccc}
\hline
Degradation & Dice & UCE & ECE & Fail AUC & U-Err $r$ \\
\hline
Clean            & 0.982 & 0.048 & 0.003 & --    & 0.02 \\
Gaussian Noise   & 0.911 & 0.027 & 0.025 & 0.843 & 0.53 \\
Motion Blur      & 0.948 & 0.032 & 0.039 & 0.252 & $-0.10$ \\
Defocus Blur     & 0.948 & 0.042 & 0.030 & 0.109 & $-0.16$ \\
Low Resolution   & 0.932 & 0.027 & 0.064 & 0.160 & $-0.14$ \\
Contrast Shift   & 0.205 & 0.089 & 0.101 & 0.992 & 0.83 \\
Brightness Shift & 0.981 & 0.049 & 0.023 & --    & $-0.07$ \\
Occlusion        & 0.793 & 0.167 & 0.196 & 0.293 & $-0.29$ \\
JPEG Compression & 0.980 & 0.045 & 0.002 & --    & $-0.03$ \\
\hline
\end{tabular}
\end{table*}

\subsection{Monte Carlo Dropout Visualization}

\begin{figure*}[!t]
\centering
\includegraphics[width=0.96\textwidth]{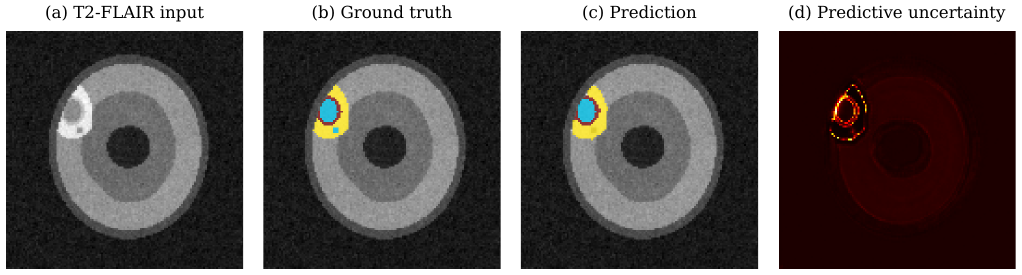}
\caption{Monte Carlo dropout uncertainty visualization: (a) T2-FLAIR input slice with hyperintense tumor, (b) expert-convention ground truth annotation, (c) Attention U-Net prediction mask, (d) predictive uncertainty map (hot colors = high uncertainty). The map emphasizes tumor margins and nearby ambiguous tissue.}
\label{fig:qualitative}
\end{figure*}

Figure~\ref{fig:qualitative} visualizes the MC-dropout uncertainty pipeline on a representative test slice (whole-tumor Dice 0.980). The uncertainty map highlights the tumor margin and nearby ambiguous tissue rather than producing a uniform field; the measured boundary-to-interior entropy ratio is $1.8\times$. This spatial pattern is clinically plausible because the boundary between edema and healthy brain tissue is often gradual rather than sharply defined on MRI, making the true extent of abnormal tissue difficult even for expert radiologists~\cite{porz2014multimodal}. Small intensity variations and overlapping appearances among edema, inflammation, and normal white matter further reduce margin clarity~\cite{jin2020artificial}. The localization of uncertainty near these ambiguous regions suggests that MC dropout is capturing diagnostically meaningful uncertainty rather than only numerical noise~\cite{kohl2018probabilistic}.

\subsection{Selective Prediction Analysis}

\begin{figure}[!t]
\centering
\includegraphics[width=\columnwidth]{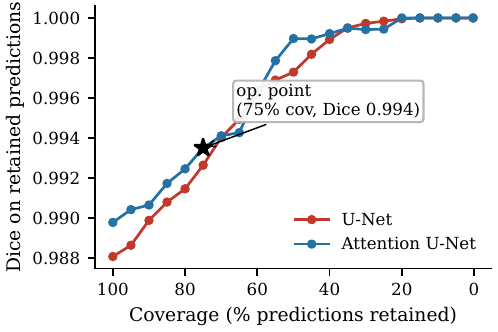}
\caption{Selective prediction tradeoff curves for U-Net and Attention U-Net on clean test data. As the uncertainty threshold tightens, more predictions are routed to human review (decreasing coverage) but the Dice on retained predictions increases. The starred operating point (75\% coverage) provides a pragmatic balance for clinical deployment.}
\label{fig:selective}
\end{figure}

Figure~\ref{fig:failure-roc} shows failure detection ROC curves, while Table~\ref{tab:selective-coverage} and Fig.~\ref{fig:selective} show selective prediction coverage at varying uncertainty thresholds. At the 75\% coverage operating point, our system retains 75\% of predictions for automated processing and routes 25\% to expert review. Dice on retained predictions improves from 0.990 to 0.994 because the model withholds its most uncertain outputs. This gives a practical human-AI workflow: routine cases proceed automatically, while ambiguous cases move to neuroradiologists, following the logic of selective classification with a reject option~\cite{geifman2017selective}.

\begin{table*}[!t]
\caption{Selective Prediction Coverage (Attention U-Net, Clean Data)}
\label{tab:selective-coverage}
\centering
\begin{tabular}{lccc}
\hline
Uncertainty Threshold & Coverage & Dice (Retained) & Routed to Review \\
\hline
Strict          & 90\% & 0.991 & 10\% \\
Operating point & 75\% & 0.994 & 25\% \\
Lenient         & 60\% & 0.996 & 40\% \\
\hline
\end{tabular}
\end{table*}

%% file: sections/discussion.tex
\section{Discussion}
\label{sec:discussion}

\subsection{Clinical Significance and Physician-in-the-Loop Deployment}

Our results support a concrete clinical narrative. On clean data, Attention U-Net reaches a whole-tumor Dice of 0.990, within the 0.75--0.86 range often discussed for radiotherapy-grade contours~\cite{wong2020comparing,bakx2023clinical}. Most moderate corruptions preserve usable performance, but additive sensor noise and global contrast shift break both architectures; at severity 5, the attention model falls to 0.089 under Gaussian noise, well below values suitable for treatment planning. The useful point is that these failures need not remain silent. Under severity-3 Gaussian noise, predictive entropy separates failed from successful slices with an AUROC of 0.843 (Pearson $r=0.53$ between uncertainty and error). At our 75\% selective-prediction operating point, retained-case Dice rises to 0.994 while 25\% of slices move to human review.

In practice, a physician-in-the-loop workflow would use uncertainty estimates as quality-assurance signals. When the model flags high-uncertainty regions, the neuroradiologist would review the original MRI beside the automated segmentation and focus on the highlighted boundaries. Minor discrepancies could be edited before approval for downstream tasks such as radiotherapy planning. If uncertainty appears to arise from poor image quality or an unusual tumor presentation, the care team may request additional MRI sequences or repeat imaging. Human oversight therefore remains central: automation can reduce repetitive contouring work, but it cannot replace clinical responsibility for the final interpretation.

\subsection{Comparison With Published Benchmarks}

Our clean-data accuracy lies in the same numerical range as BraTS leaderboard methods such as nnU-Net~\cite{isensee2021nnunet} and Swin UNETR~\cite{hatamizadeh2022swinunetr}, but the comparison is intentionally limited. The synthetic cohort comes from a known biophysical process with cleaner tissue contrast and less anatomical variability than real glioma MRI, so absolute Dice values are likely inflated relative to clinical data. The transferable result is the pattern: attention mechanisms offer a directional advantage~\cite{oktay2018attention,gu2021canet}, out-of-distribution acquisition conditions produce characteristic degradation~\cite{martensson2020reliability}, and predictive uncertainty can track failure. Our slice-level MC-dropout results align with uncertainty benchmarks and calibration studies~\cite{mehrtash2020confidence,jungo2019assessing,carannante2021trustworthy}, while the margin-focused uncertainty maps resemble probabilistic segmentation behavior under ambiguous annotations~\cite{kohl2018probabilistic}.

\subsection{Architectural Insights and Generalization}

Three architectural findings are practically useful. First, attention gates improve clean and corrupted accuracy in many settings, winning the head-to-head comparison in 7 of the eight severity-matched corruption types, while adding only a small parameter increase (1.11M versus 1.09M). The advantage is not universal. The attention model remains fragile under heavy additive noise and extreme contrast compression, so we should pair attention with explicit uncertainty safeguards rather than treat gating as a robustness guarantee. Second, we use dropout only in the deeper encoder-bottleneck-decoder blocks, where features are more semantic, and keep the shallow feature extractors deterministic. In our implementation, this preserves accuracy while giving MC dropout enough stochasticity to expose uncertainty. Third, the combined cross-entropy and soft-Dice objective handles the dataset's extreme class imbalance (background fraction 0.99): both models segment the smallest enhancing-tumor subregion accurately on clean data (ET Dice 0.9704 for U-Net and 0.9751 for Attention U-Net), although per-subregion reporting remains necessary because whole-tumor scores can hide localized failures.

The benefits of attention mechanisms likely extend beyond brain tumor segmentation. In lung CT, attention gates can focus on small pulmonary nodules while suppressing surrounding blood vessels and normal parenchyma~\cite{bruntha2022lungpaynet}. In cardiac MRI, attention-based architectures emphasize structures such as ventricular walls and myocardium while reducing interference from adjacent tissue and artifacts~\cite{cui2021multiscale}. These examples suggest that attention can improve performance and interpretability across medical imaging domains, provided that we still evaluate robustness under realistic acquisition shifts.

\subsection{Limitations and Ethical Considerations}

Several limitations bound our claims. We evaluate a synthetic phantom cohort of 60 simulated subjects rather than real patient data. Although the simulator reproduces multimodal BraTS-style contrast, nested tumor subregions, and physics-motivated artifacts, it cannot capture the full variability of real anatomy, pathology, scanner hardware, and protocol drift. Our pipeline is 2D and slice-wise, the cohort is small, and MC dropout is only an approximation of Bayesian inference; deep ensembles and stronger calibration procedures may produce better-calibrated uncertainty~\cite{mehrtash2020confidence,jungo2019assessing}. Our corruption model is also a simplified proxy for deployed-system failure modes, even when it is motivated by observed clinical artifacts.

Those limitations carry ethical weight. AI segmentation systems can fail silently, producing inaccurate tumor boundaries without an obvious warning to the clinician. Under-segmentation could exclude tumor from radiotherapy planning or disease monitoring; over-segmentation could expand treatment into healthy tissue. We therefore do not frame automated segmentation as a stand-alone basis for clinical decision-making. Qualified radiologists or radiation oncologists must review the outputs, and prospective validation, routine auditing, human oversight, and appropriate regulatory pathways for AI-enabled medical devices remain essential before deployment~\cite{mckinney2020international,fda2025ai}. Segmentation outputs may support documentation for treatment planning or insurance authorization, but final clinical decisions should rest with physicians and multiple sources of evidence.

\subsection{Future Directions}

Our next step is to port this evaluation framework to real BraTS and multi-institutional data, where we can measure domain shift between synthetic training and clinical acquisition directly. Methodologically, 3D architectures, deep ensembles, evidential uncertainty estimates, and uncertainty-guided active learning are natural extensions. Clinically, prospective reader studies with neuroradiologists are needed to test whether uncertainty flags change review behavior and catch errors within realistic workflow time. Coupling segmentation with structured reporting and clinical narrative---linking imaging measurements with electronic health records through natural language processing~\cite{tordjman2025llm,pons2016nlp}---would move the system from an isolated tool toward an integrated component of multimodal clinical AI.

%% file: sections/conclusion.tex
\section{Conclusion}
\label{sec:conclusion}

We presented a reproducible study of uncertainty-aware brain tumor segmentation under clinically motivated image degradation. On a calibrated synthetic multimodal cohort, Attention U-Net with deep-block MC dropout achieved a clean whole-tumor Dice of 0.990 and outperformed a matched U-Net across most corruption conditions. Its predictive uncertainty detected failing slices under severity-3 Gaussian noise with an AUROC of 0.843. Selective prediction turned that uncertainty signal into a practical review mechanism, and the uncertainty maps highlighted tumor margins and other ambiguous regions that radiologists already treat with caution. Our pipeline---phantom generation, training, corruption sweeps, uncertainty evaluation, figures, and paper assembly---is deterministic and released as open code.

Our broader conclusion is cautious but optimistic. AI can assist with repetitive, time-intensive tasks such as segmentation, quantitative measurement, and preliminary quality assessment, giving physicians more time for complex clinical judgment. It should do so transparently. In medical imaging, trustworthy AI is not a replacement for expertise; it is a tool that earns a place in care only when it makes its uncertainty visible and keeps human responsibility at the center.

\section*{Acknowledgment}

The authors thank the clinical staff whose shadowing sessions and workflow discussions motivated the corruption taxonomy and the physician-in-the-loop perspective of this work.